# Machine Learning Based Early Fire Detection System using a Low-Cost Drone

Ayşegül Yanık[1], Mehmet Serdar Güzel[2*] , Mertkan Yanık[3] and  Erkan Bostancı[4]

[1,2,3,4] Computer Engineering Department of Ankara University
**Corresponding author e-mail:** mguzel@ankara.edu.tr

**Abstract:** This paper proposes a new machine learning based system for forest fire earlier detection in a low-cost and accurate manner. Accordingly, it is aimed to bring a new and definite perspective to visual detection in forest fires. A drone is constructed for this purpose. The microcontroller in the system has been programmed by training with deep learning methods, and the unmanned aerial vehicle has been given the ability to recognize the smoke, the earliest sign of fire detection. The common problem in the prevalent algorithms used in fire detection is the high false alarm and overlook rates. Confirming the result obtained from the visualization with an additional supervision stage will increase the reliability of the system as well as guarantee the accuracy of the result. Due to the mobile vision ability of the unmanned aerial vehicle, the data can be controlled from any point of view clearly and continuously. System performance are validated by conducting experiments in both simulation and physical environments.

## 1. Introduction

The perspective based on visual indications relating to early detection systems in forest fires. In this period where the number of systems developed by utilizing unmanned aerial technology is increasing day by day, unmanned aerial vehicles will be used to achieve the targets of minimizing the destruction of our forests which are the lungs of the world and optimizing the usage of workforce and time resources [1, 2, 3]. As a result of the application carried out in line with the subject of this paper, it is proposed that the system based on the detection of smoke image with unmanned aerial vehicle can provide a great benefit in reducing the error rate occurring in fire detection. The microprocessor in the system has been trained with deep learning methods and has been given the ability to recognize smoke image, which is the earliest sign of fire diagnosis. The most fundamental problem in the common algorithms used in fire detection is the high level of false alarm and overlook rate [4,5].Confirming the result obtained from the detection and defining an additional proof will increase the reliability of the system as well as the accuracy. Since the drones provide a mobile vision, the point of view can be controlled by the ground station can manipulate it for the sake of the accuracy of the result. The application developed in line with the subject of the paper was implemented in both simulation and physical environments and the advantages of early fire detection system and analysis results are discussed in the conclusion section of the article.

*1.1. Motivation*

Once the clues of the fire is inspected in a real time process, it has obvious distinguishing features both in terms of motion, color spectrum and textural structure. In this way, it can be easily separated from other natural components in the sky and forest by means of various filtering, edge detection and color recognition tools. However, it is too late to intervene when the fire and flame become visible. For this reason, based on the smoke data, which is the earliest sign of the fire, convincing the fire diagnosis will pave the way for a faster reaction. Nevertheless, the results obtained by processing image or video input by image processing algorithms in today's developed applications contain high errors. In the event of a fire, no signal is given or the system goes into an alarm state when there is no fire. These conditions should be minimized for early and on-site intervention. At this stage, a control system patrolling with unmanned aerial vehicles will be functional in terms of providing fire and providing a clearer view [7, 8]. The choice of a microprocessor controlled by an electrically powered microprocessor from the power component of unmanned aerial vehicles will save time in the assessment of the possibility of fire using the "TensorFlow library" of the microprocessor in the vehicle without the data received from the vehicle being sent to the control station for artificial smoke detection. The main objective of this study is to minimize the rate of false reports and omissions and to optimize the process. Designed Drone is illustrated in Figure 1 As well as Bottom and Top body designs of the drone is given in Appendixes A and B Overall section 2 involves materials and method whereas sections 3 will followed by the experimental results section and the paper is finally concluded.

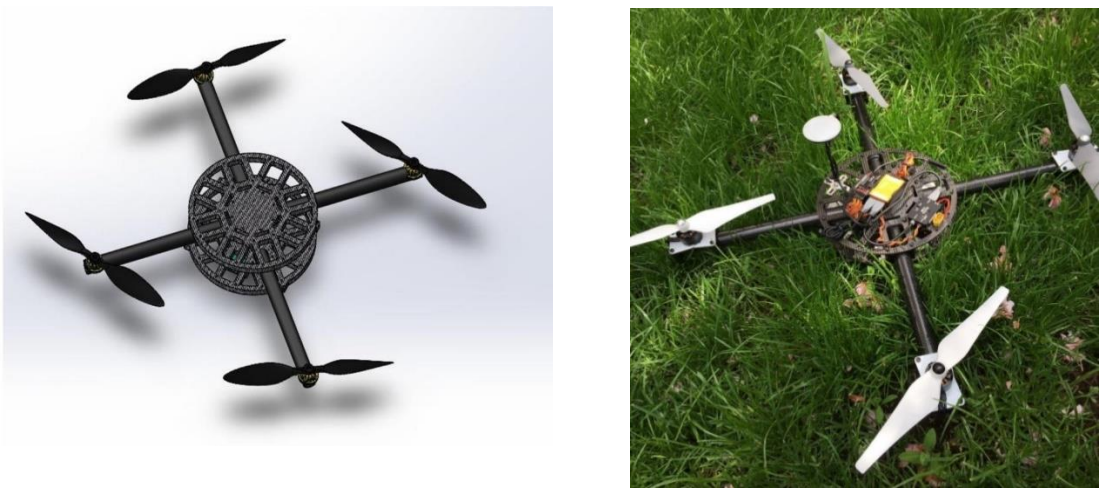

**Figure 1.** 3-D Solidworks model of the designed drone (left), the final version of the drone (right)

## 2. Materials and Methods

Forest fire is a natural disaster and its havoc can be significantly reduced in case of early detection and early intervention. Despite the superior efforts of the firefighters, massive diffusion of fire is unavoidable for some reasons such as traffic, late or false alarms, in case of where resides hard to intervention. In the case of forest fires the location and the time required to reach are the most important obstacle to be overcome. Because forest fires have the fastest spreading speed and the most noticeable type of fire [9,10]. Due to the fact that forest fires that occur frequently, green areas in our world are getting smaller each day. Forest fires have been intensified in some provinces. It is known that helicopters are involved in firefighting besides water tankers and pumper vehicles and so faster results have been obtained from land vehicles [11].

The objective of this study is to minimize the rate of error in the results obtained in detecting the forest fires by the ability of the unmanned aerial vehicles to view and move. The Raspberry Pi device (see Figures 2 and 3) is a powerful and high capacity device used in many robotic and IoT projects, offering the programmer a mobile production environment. The first step in making the Raspberry Pi minicomputer usable with unmanned aerial vehicles will be to connect the device to the system, which can be fed from the internal battery of the unmanned aerial vehicle. Otherwise, other methods that can be used, such as feeding the device with a powerbank to be fixed on the unmanned aerial vehicle, are not a desired feature since it will increase the weight of the vehicle.

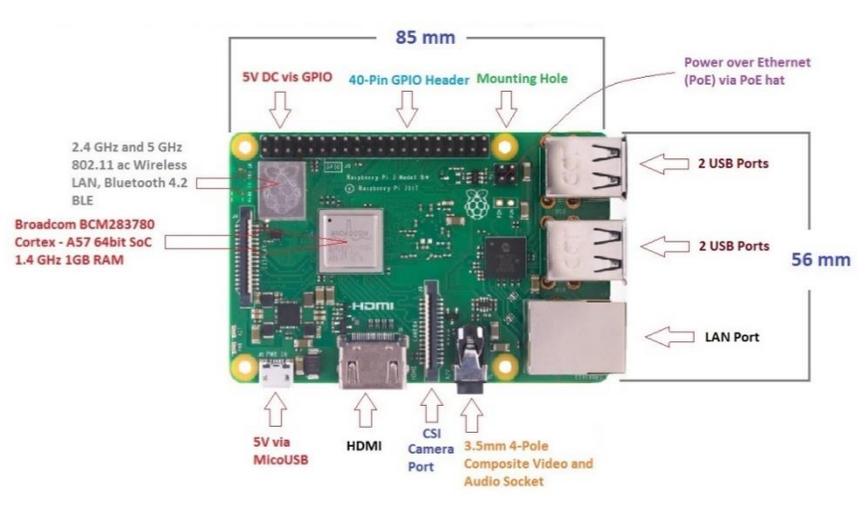

**Figure 2.** 3-D Configuration of Raspberry Pi device

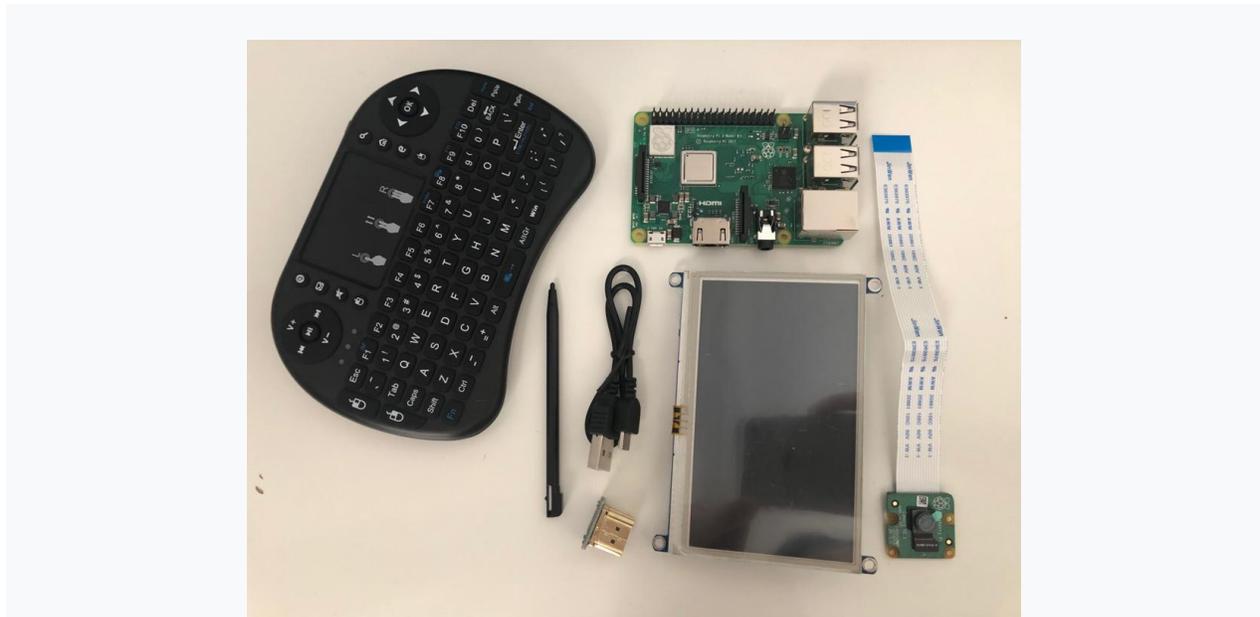

**Figure 3.** 3-D Raspberry Pi device and other devices used in this project

The current and voltage values of the electricity required from the LiPo battery of the Raspberry Pi are too high that it prevents feeding the device from the vehicle with trip wiring. For Raspberry, this regulator is placed between the power distributor and the inlet slot to overcome this obstacle. A converter battery adjuster circuit for this, whereas, the black pin end pin ends of the BEC circuit connect to the ground (ground) end of the power distributor, the red pin end of the circuit to the voltage-end of the distributor, and the triple end of the rechargeable circuit to the Raspberry Pi 2.4 and 6 function pin inputs. While designing a frame for the four-engine rotary wing unmanned aerial vehicle to be used in the project, a perforated model made of carbon fiber material was preferred by considering the aerodynamic effect and the center of gravity distribution. When looking at the drone designs, the reason for choosing the H-type frame can be seen as the potential of the drones with this type of frame to perform a more robust roll while performing load-bearing tasks due to the reliable mass base and arms extending apart from the frame. H-type drones can recover more easily after taking sharp bends, and it can be predicted that the battery placed in the center of the hull will last longer as it will take less damage in possible drops. However, in the original design of the vehicle used in the project, it is planned to place electronic components in the sandwich model on the coreless body. Thus, since the electronics and battery will be located in a sheltered compartment during trials and flight training, the second of the two main differences between the X and H type cases remains ineffective. The fastest move to be made during the fulfillment of the project requests will be the movement to the left and right. Considering this requirement, the most obvious difference that makes the H frame reasonable when the X and Y frames are compared will also lose its importance. Generally, type X frame is used in racing drones. Considering that the fast course of the vehicle to follow will provide a wider control, the advantages of using the X-type drone in racing vehicles played a role in the choice of the X-type body in the body design of the project vehicle. These advantages are; more balanced distribution of the center of gravity, the mechanisms involved in drone moves are in a symmetrical co-operating principle, and the pilot can be more easily orientated to the vehicle control commands in control and control. In the project vehicle, the distance of the engines from the body / center arm length was designed as 30 cm. The four-engine rotary wing unmanned aerial vehicles have a well-established place with distinct advantages among lightweight multicopters. The most important advantage is that the unstable drone design minimizes the balance handicap due to the motor / wing symmetry compared to the fixed wings. Each engine to be added will add weight to the design, but at the same time, since the increase in engine power will directly increase the handling, acceleration capacity (torque) and performance of the drone, the four-engine design will balance this contrast well, producing the most appropriate results. The electronic speed control unit, engine and propeller components make the drone propulsion system. Speed control in brushless motors can be done with constant torque, since there is no friction, the motor; it does not arc, dust, warms up and works at high efficiency. In the design of the vehicle, it was found appropriate to use four 770 KV short shaft avroto brushless motors. In each engine, four propellers made of high-strength plastic / nylon or carbon fiber nylon material will be used, taking into account the estimated weight of the project vehicle and the risks that may arise in possible collision. Instead of 3 blades of the propeller to be selected, it has been determined as 2 blades, inclination angle of 4.5 "(11.43 cm) and propeller diameter of 10" (25.4 cm). This selection was based on peer racing drones. When calculating the thrust value of the unmanned aircraft system, d symbolizes the propeller diameter in inches and is included in the formula as 10 "(25.4 cm), which is a known variable. RPM value represents the numerical value of the propeller in one minute. The pitch value is represents the value of the road traveled in a propeller lap in inches, and V0 represents the forward propeller speed in m / s unit. In the forward dynamic propulsion equation, the test data of the Avroto M2814-11S engines used in the vehicle was used when calculating the thrust value [11]. The fact that the vehicle has a unique design is one of the main objectives and achievements in the project. In order to carry the sub-systems of the vehicle (Raspberry Pi, Camera Module, Landing Gear, Battery, 4 ESC etc.) in a balanced and sheltered way, the model formed by the parallel placement of two empty plates called the sandwich model has been found appropriate. Taking into account the aerodynamic elements, the sandwich model of the vehicle is supported by the perforated body design (See Figure 1). After the vehicle was created, flight attempts were made in two ways by mounting the battery above and below the case. The working principle of the load carrying mechanism in unmanned aerial vehicles with rotary wings is based on the force of the air produced by the thrust units downward. Thus, any design in which propellers interact with each other or another object negatively affects the stability and performance of the vehicle. After this was observed in the test results, the batteries were placed under the body. During flight tests, no losses were made except for the propeller.

*2.1. Software Design*

Raspbian operating system is installed on the device, which has been developed and can run many programs smoothly and quickly. OpenCV is suitable for the use of image processing and also installed into the device. Besides, Tensor Flow library, a leading machine learning library, is also installed for object detection problem. It is possible to install and make programs with standard Linux commands through the terminal command line after the installation of the operating system on the microcontroller. Python 3.6 language is utilized to bring artificial learning skills into the tool. In the model used to distinguish the smoke image from the background elements, several post processing steps are carried out involving, background separation, color and edge modeling, and data normalization. With the help of built-in methods, functions and defined values in the libraries used in this steps, the steps to be taken during the calling and use of the model after the training phase are carried out in a cost-free manner and in order to prevent the processor from being exhausted. In the subheadings of the Materials and Methods section, the installation and start-up steps of the application, the identification and training of the model, and the process of installing the model on a newly purchased Raspberry are included. As aforementioned above, Raspberry Pi is supported by OpenCV and Tensorflow libraries, integrated with a self-made drone so as to detect forest fires. Tensorflow is a library used to train the system for the object to be distinguished in object recognition applications. In the Tensorflow model, the distinguishing properties of the object are extracted using a data type called tensor. In terms of the Tensorflow library, tensor is a multidimensional data structure consisting of primitive (integer, float) data contained within a multidimensional array. It serves to keep the difference and similarity mathematically determined when certain regions of the images in the stream are specified to contain the same object. The Tensorflow diagram aims to identify the factors that are effective in the most accurate classification of test images and to produce a property map for the objects in the model, with improvements and training phases. When the object recognition algorithm is running, a series of computational operations, which are made up of nodes and can be represented by a flow diagram, are executed when the object to be distinguished is identified in the model. Generally, each node takes a tensor, calculates it and outputs a tensor. In learning models, it is aimed to get different and improved outputs with the same inputs. In order to achieve this, tensors in the flow are updated and iteration is maintained by using Tensorflow variables, operations, placeholders and constants. Improvement updates in machine learning cannot be done manually, there is an API called loss function, which will minimize the instantaneous value of the function used to measure the difference between the estimated success value of the model and the success of correlation value is fed to that object as a similar relationship is encountered in the region containing the object in the other pictures. The training pictures in the batch, the specific iteration unit, are finally compared with the test data and the model is updated and optimized according to the difference between the classification and the class to which it actually belongs. The statistics and results of the mathematical evaluations that the trained system is trained to give fast and low error rates are given in this section. The ssdlite mobilenet model, which is a sub-model of the supervised deep learning school called Model Zoo, which contains pre-educated models in the Tensorflow library, is taken as an example for training in this study [12,13]. There is a speed-precision swap that creates a kind of balance between sub-models. The criterion to be considered when choosing a model is the purpose for which the model will be used. Because in programming there is always a trade-off balance that makes it necessary to waive certain things while improving certain things. In the real-time object recognition application, the speed of the model is in the foreground, while the accuracy rate remains in the background. Essentially, the models, trained with COCO [12,14] training set, were investigated.

According to which, the "ssdlite mobilenet" model, most likely to give the best results when working on real time smoke data on Raspberry, is adapted for this study. An example smoke detection within the developed software system is given in Figure 4.

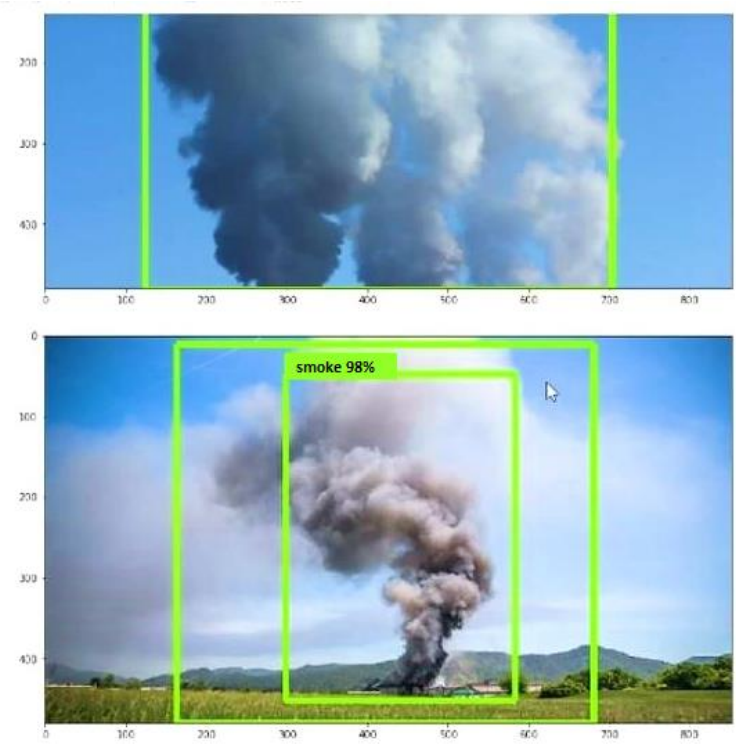

**Figure 4.** An Example Smoke Detection with (%98) probability

For the training process, 280 images, representing smoke, most of those obtained from google whereas some of them are generated in physical environments. Consequently, those data are employed the training phase of the model . Change in the loss functions is illustrated in Figure 5.  Figure 6 illustrates images obtained from datasets, whereas Figure 7 illustrates an example data generated by the authors by using the developed drone.

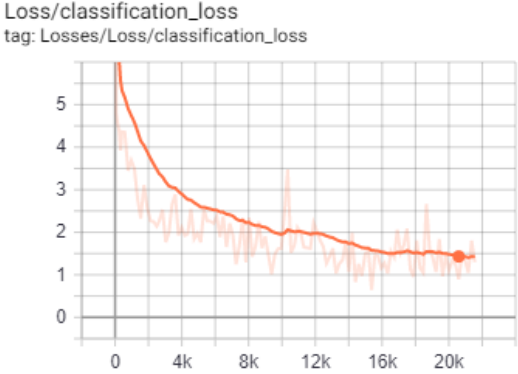

Figure 5. Loss Function variation (y-axis) during the iterations (x-axis) (%90 of data for training)

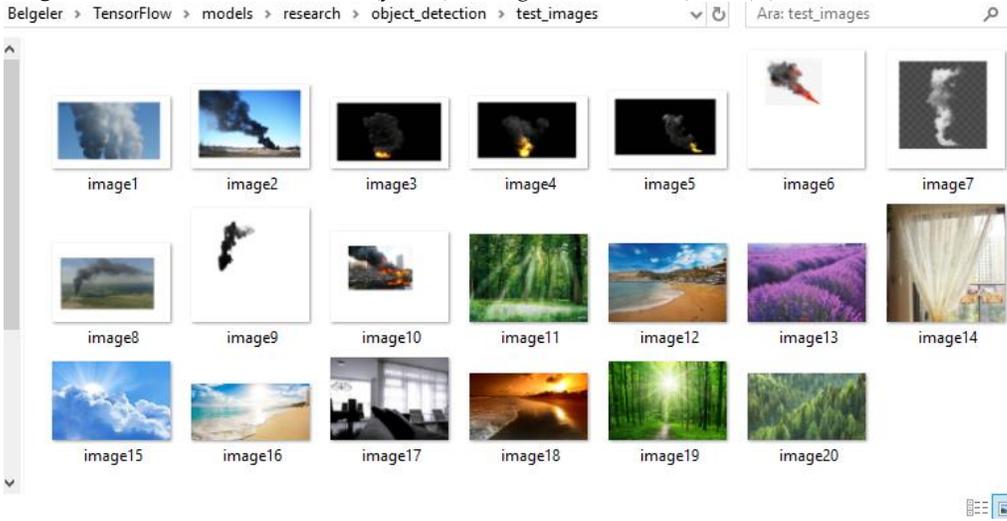

Figure 6. Dataset obtained from Google and corresponding OA datasets

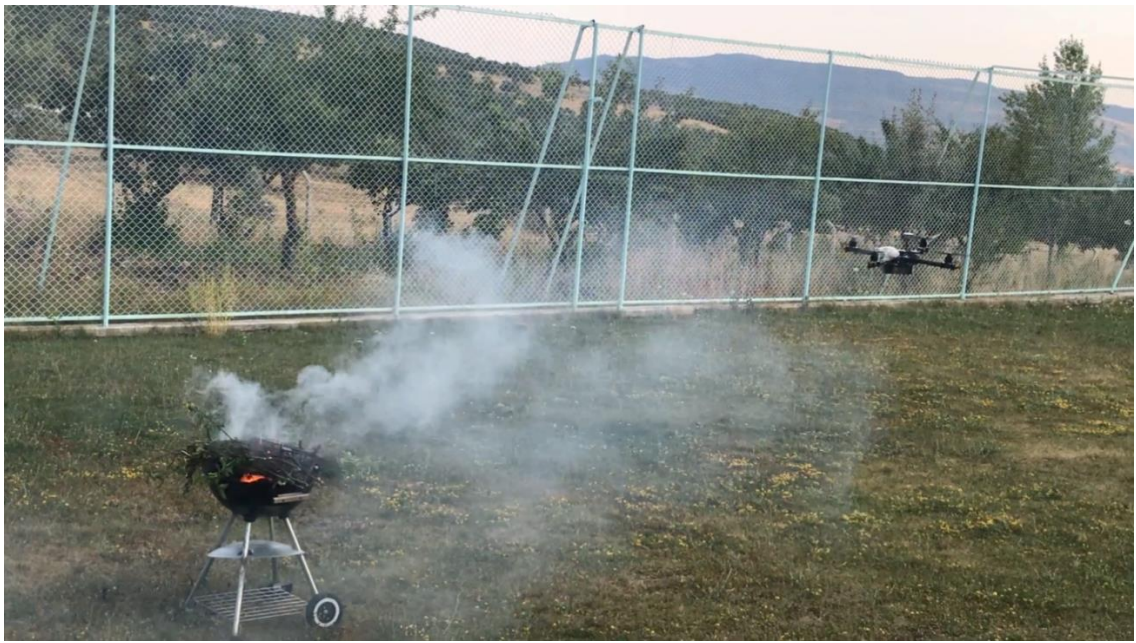

Figure 7. An example data generated by the author using the developed drone

## 3. Results

When evaluating the loss graphs, it would not be correct to call the model with the lowest loss value that we obtained the most successful model. This can only show that that model can be trained more quickly than other models. The training was ended in different models and different iterations in each model. For instance, Figure 5 illustrates a model employs %90 of data for training. Comparing the results to the point where each model reaches the optimum level, comparing the results and comparing the loss values at the end of the trainings can be considered as a relatively accurate comparison parameter. There are two versions of Tensorflow running on CPU and GPU. The model is trained in two ways. The same model working on the same data was able to reach the point where the training process carried out on the CPU came within 10 hours 32 minutes 54 seconds, on the GPU in 1 hour 39 minutes 47 seconds. In this case, it was observed that the model was trained 6.34 times faster compared to the CPU when the deep learning algorithms carried out on visual data were carried out on the GPU. A number of analysis procedures were carried out on four models run on the same control group to choose between the different percentage of training and test data dropped and four different models trained. Estimation success of the models in different respects was measured and models were compared to each other and one of them was preferred to be used. There are four models are defined based on the distribution of the train and test data.

**Table 1.** Definition of Models used in the experimental part

| Model Name | Training Data (%) | Test Data (%) |
|---|---|---|
| M1 | 60 | 40 |
| M2 | 70 | 30 |
| M3 | 80 | 20 |
| M4 | 90 | 10 |

There are some standard evaluation parameters used to measure the overall performance of deep learning models. These are total accuracy, positive interpretation power, negative interpretation power, sensitivity, F-rating and specificity. When calculating these values, DP: true-positive, DN: true-negative, YP: false-positive and YN: false-negative FN: false-negative is used [16-18]. The true-positive expression refers to the number of operations in the sample space where the work carried out in the dissertation project, in which the model also contains smoke for visual data that actually contains smoke. The number of true-negative indicates the number of visual data that the model classifies as smoke-containing. On the one hand, the false-positive does not actually contain smoke; on the other hand, the false-negative characterizes the number of visual data classified by the model as though it actually contains smoke, though it does not contain smoke. The fact that the true-positive and true-negative values are high shows that the model produces realistic results. Models with a large number of false-positive numbers are likely to be triggered by the fire department in the absence of fire, and models with a large number of false-negative numbers that can lead to negligence by not seeing the fire. In this case, conducting such a numerical analysis before deciding on the model to be used is of great importance for the efficiency of the project. For all four models, total accuracy, positive interpretation power, negative interpretation power, sensitivity, F-evaluation and specificity values of the models were calculated based on the results of the estimation values table prepared using the values recorded in the output table. Total accuracy is the most intuitive and basic of the evaluation criteria. Those equations are given below:

$$accuracy = \frac{TP+TN}{TP+FP+TN+FN} \qquad (1)$$

$$precision = \frac{TP}{TP+FP} \qquad (2)$$

$$recall = \frac{TP}{TP+FN} \qquad (3)$$

$$Fmeasure = 2x \frac{precision \times recall}{precision + recall} \tag{4}$$

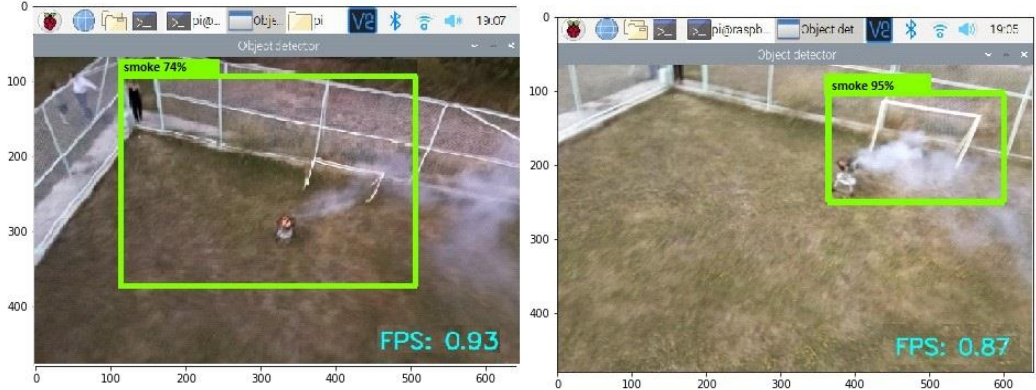

**Figure 8.** Some test results obtained from the physical environment

**Table 2.** Performances evaluation of training models used in the experimental part

| Model | $M_1$ | $M_2$ | $M_3$ | $M_4$ |
|---|---|---|---|---|
| Accuracy | 0.55 | 0.55 | 0.85 | 0.7 |
| Precision | 0.7 | 0.5 | 0.7 | 0.7 |
| Recall | 0.4 | 0.3 | 0.8 | 0.6 |
| FScore | 0.51 | 0.38 | 0.75 | 0.64 |

A number of analysis procedures were carried out on four models run on the same control group to choose between the different percentage of training and test data dropped and four different models trained. Estimation success of the models in different respects was measured and models were compared to each other and one of them was preferred to be used. The preferred model is the third model (M3), trained with 70% training and 30% test data. Table 2 demonstrates the overall results of the corresponding physical experiments. According to the which, Model 3 is determined as the best model for all evaluation criterias. For the physical experiment M3 model was loaded on Raspberry Pi and test studies were carried out. In order to carry out the flight test in accordance with the rules, near the Kızılcahamam Soguksu Mili Park, a remote and sheltered area was preferred, and the test flight and related shots were carried out using daylight. As a smoke source, it was ignited with brazed gel in footed enamel brazier, and some plastic waste material was added for the brushing and bulking of smoke (See Figure 7). Figure 8 illustrates experimental result both validating the performance of the drone and its smoke recognition capacity. During the test flight of the system (See Figure 7), images of smoke and vehicle were taken both from inside Raspberry (See Figure 7) and from outside. Since the vehicle does not have a vision system (FPV, First Person View) that can help the pilot to control the vehicle more easily, it has been very difficult to center the smoke in the images taken from the vehicle. For this reason, it is recommended to use a visual support mechanism while driving in similar studies. In addition, due to vibration obtained images are blurry. However, even though the smoke recognition model worked flawlessly, negligible localization errors were observed and the similarity rate was not successful in each smoke input, the test flights was successfully completed.

## 5. Conclusions

This paper introduces a new low-cost drone equipped with image processing and object detection abilities for smoke and fire recognition tasks in forests. The statistics and results of the mathematical evaluations that the system is trained to give fast and low error rates are given in this section. The ssdlite mobilenet model, which is a sub-model of the supervised deep learning model called Model Zoo, which contains pre-educated models in the Tensorflow library, is taken as an example to be educated in the thesis project. There is a speed-precision trade-off that creates a kind of balance between sub-models. The criterion to be considered when selecting a model is the purpose for which the model will be used. Because in programming there is always a trade-off balance that makes it necessary to waive certain things while improving certain things. In the real-time object recognition application, the speed of the model is in the foreground and the accuracy rate remains in the minor preferences. In the project, the models that were trained with COCO training set were investigated and the ssdlite mobilenet model which was most likely to give the best results when working on real time smoke data on Raspberry was modified. When evaluating the suitability of the model used, it should be taken into consideration that each model cannot work with the same speed and accuracy rate in each data set. A model based on the dimension of colors cannot achieve the success of distinguishing the objects of color that distinguishes itself from the background on objects that are very close to the background. The preferred model has successfully performed its task. Based on these results, it was considered appropriate to use the drone and the early fire detection system for the detection of forest fires. In order to improve the system, it is possible to increase the maximum flight time of the unmanned aircraft so that the area that can be controlled in a single flight can be expanded. In this eco-friendly project, the system can be supplied from convertible energy sources such as solar energy instead of LiPo battery. In addition, in cases where smoke is detected in the unmanned aircraft, it is possible to estimate how far the smoke is from the vehicle by means of monocular depth detection algorithms, and this improvement enables the location of the fire to be reported to the ground station. In addition to the smoke recognition application, the same model can be trained to detect flame and fire visuals, and a downward camera can be added to the unmanned aerial vehicle to perform a downward fire / flame scan on the vehicle suspended above the smoke detection zone. This will enhance the accuracy of the project outcome. Hoping that the study will shed light on similar scientific studies.

**Acknowledgments:** Some part of this study was published in the MSc thesis entitled with "Visual Based Early Fire Detection System with Unmanned Aerial Vehicles, 2019" of the first author.

**Conflicts of Interest:** "The authors declare no conflict of interest." the results".

**Appendix A**

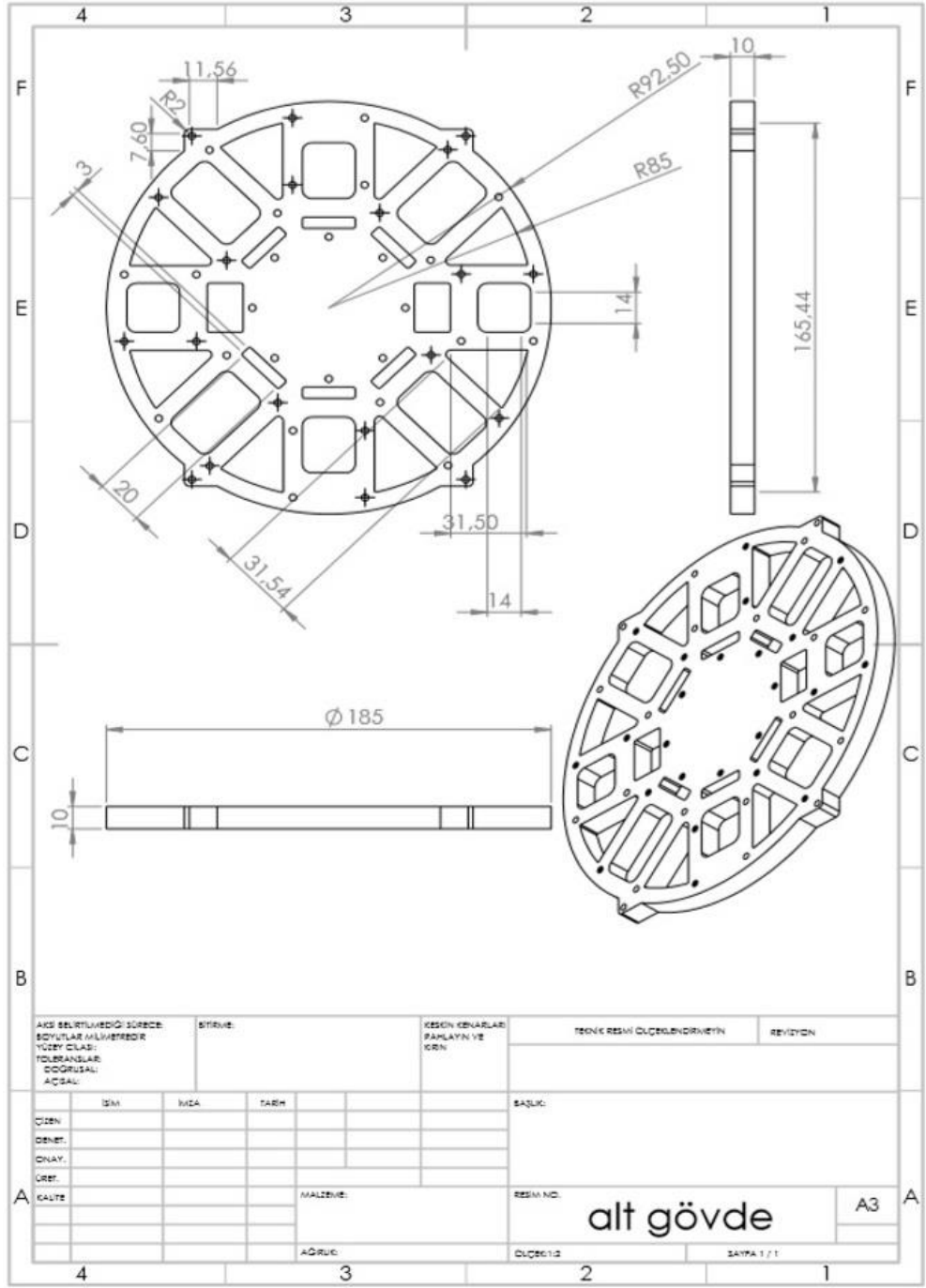

**Figure A1.** Bottom body design of the developed Drone [13]

**Appendix B**

**Figure A2.** Top body design of the developed Drone [13]

**References**

[1] Boroujeni, N. S. (2019). Monocular vision system for unmanned aerial vehicles. doi:10.22215/etd/2013-07242.

[2] Akyurek, S., Yılmaz, M.A., and Taskıran, M. (2012)., "İnsansız Hava Araçları(Muharebe Alanında ve Terörle Mücadelede Devrimsel Dönüşüm", technical report No: 53, Ankara, TR.

[3] Chen, T., Wu, P., & Chiou, Y. (2014). An early fire-detection method based on image processing. *2004 International Conference on Image Processing (ICIP04)*, Singapore.

[4] Gao, Y., & Cheng, P. (2019). Forest Fire Smoke Detection Based on Visual Smoke Root and Diffusion Model.*Fire Technology*. doi:10.1007/s10694-019-00831-x.

[5] Gao, Y., & Cheng, P. (2019). Forest Fire Smoke Detection Based on Visual Smoke Root and Diffusion Model.*Fire Technology*. doi:10.1007/s10694-019-00831-x.

[6] Wang, T., Liu, Y., & Xie, Z. (2011). Flutter Analysis Based Video Smoke Detection. *Journal of Electronics & Information Technology*,33(5), 1024-1029. doi:10.3724/sp.j.1146.2010.00912.

[7] Lin et. al, Smoke detection in video sequences based on dynamic texture using volume local binary patterns. (2017). *KSII Transactions on Internet and Information Systems*, (11). doi:10.3837/tiis.2017.11.019.

[8] Lin et. al, Wang, Z., Wang, Z., Zhang, H., & Guo, X. (2017). A Novel Fire Detection Approach Based on CNN-SVM Using Tensorflow. *Intelligent Computing Methodologies Lecture Notes in Computer Science*,682-693.

[9] Shaqura, M., & Shamma, J. S. (2017). An Automated Quadcopter CAD based Design and Modeling Platform using Solidworks API and Smart Dynamic Assembly. *Proceedings of the 14th International Conference on Informatics in Control, Automation and Robotics*. doi:10.5220/0006438601220131.

[10] O. T. Cetinkaya, S. Sandal, E. Bostancı, M. S. Güzel, M. Osmanoğlu and N. Kanwal, "A Fuzzy Rule Based Visual Human Tracking System for Drones," *2019 4th International Conference on Computer Science and Engineering (UBMK)*, Samsun, Turkey, 2019, pp. 1-6, doi: 10.1109/UBMK.2019.8907104.

[11] M. Unal, E. Bostanci, E. Sertalp, M. S. Guzel and N. Kanwal, "Geo-location Based Augmented Reality Application For Cultural Heritage Using Drones," *2018 2nd International Symposium on Multidisciplinary Studies and Innovative Technologies (ISMSIT)*, Ankara, 2018, pp. 1-4.

[12] K. He, G. Gkioxari, P. Dollár and R. Girshick, "Mask R-CNN," in *IEEE Transactions on Pattern Analysis and Machine Intelligence*, vol. 42, no. 2, pp. 386-397, 2020,.doi: 10.1109/TPAMI.2018.2844175.

[13] Gad, A. F. (2018). Tensorflow Recognition Application. *Practical Computer Vision Applications Using Deep Learning with CNNs*,229-294. doi:10.1007/978-1-4842-4167-7_6.

[14] Yilmaz, A.A., Guzel, M.S., Askerbeyli, I., et al, 2018, 'A vehicle detection approach using deep learning methodologies', arXiv:1804.00429.

[15] Demirtaş (Yanık), Ayşegül, Visual Based Early Fire Detection System with Unmanned Aerial Vehicles , MSc thesis, 2019, Ankara University , TR

[16] A. A. Yilmaz, M. S. Guzel, E. Bostanci and I. Askerzade, "A Novel Action Recognition Framework Based on Deep-Learning and Genetic Algorithms," in *IEEE Access*, vol. 8, pp. 100631-100644, 2020, doi: 10.1109/ACCESS.2020.2997962.

[17] Karim, A.M.; Kaya, H.; Güzel, M.S.; Tolun, M.R.; Çelebi, F.V.; Mishra, A. A Novel Framework Using Deep Auto-Encoders Based Linear Model for Data Classification. *Sensors* **2020**, *20*, 6378. https://doi.org/10.3390/s20216378

[18] Güzel MS, Kara M, Beyazkılıç MS. An adaptive framework for mobile robot navigation. *Adaptive Behavior*. 2017;25(1):30-39. doi:[10.1177/1059712316685875](10.1177/1059712316685875)